\pdfoutput=1

\documentclass[11pt]{article}
\usepackage{acl}

\usepackage{times}
\usepackage{latexsym}

\usepackage[T1]{fontenc}

\usepackage[utf8]{inputenc}

\usepackage{microtype}

\usepackage{inconsolata}

\usepackage{todonotes}

\newcommand{\yy}[1]{{\color{black}{#1}}}
\newcommand{\yynew}[1]{{\color{black}{#1}}}

\newcommand{\modify}[1]{\textcolor{black}{#1}}
\newcommand{\origin}[1]{\textcolor{black}{#1}}
\newcommand{\revise}[1]{{\textcolor{black}{#1}}}

\newcommand{\RNum}[1]{\uppercase\expandafter{\romannumeral #1\relax}}
\usepackage{graphicx}
\usepackage{multirow}
\usepackage{amsfonts}
\usepackage{booktabs}
\usepackage{subcaption}
\usepackage{amsmath}
\usepackage{cleveref}

\title{
ATG: Benchmarking Automated Theorem Generation for Generative Language Models
}

\author{%
  \space Xiaohan Lin$^{1}$\space\space\space
  Qingxing Cao$^{1}$\thanks{~~Corresponding author.},\space\space\space
  Yinya Huang$^{2}$,\space\space\space
  Zhicheng Yang$^{3}$,
  \\
  \textbf{
  Zhengying Liu$^{4}$,\space\space\space
  Zhenguo Li$^{4}$,\space\space\space
  Xiaodan Liang$^{1}$$^{5}$$^{*}$,\space\space\space 
  }
  \\
  $^{1}$\normalfont{Shenzhen Campus of Sun Yat-Sen University}\quad 
  $^{2}$\normalfont{City University of Hong Kong}\quad  \\
  $^{3}$\normalfont{The Hong Kong University of Science and Technology (Guangzhou)}\quad  \\
  $^{4}$\normalfont{Huawei Noah’s Ark Lab}\quad
  $^{5}$\normalfont{DarkMatter AI Research}\quad
  \vspace{0.15cm} \\
  {\tt\small \{linxh55,  caoqx\}@mail2.sysu.edu.cn}, {\tt\small \{yinya.huang\}@hotmail.com}\\
  {\tt\small\{yangzhch6, xdliang328\}@gmail.com}, {\tt\small\{liuzhengying2, Li.Zhenguo\}@huawei.com} \AND 
}
\begin{document}
\maketitle
\begin{abstract}
Humans can develop new theorems to explore broader and more complex mathematical results.
While current generative language models (LMs) have achieved significant improvement in automatically proving theorems, their ability to generate new or reusable theorems is still under-explored. Without the new theorems, current LMs struggle to prove harder theorems that are distant from the given hypotheses with the exponentially growing search space.
%
Therefore, this paper proposes an \textbf{\underline{A}}utomated \textbf{\underline{T}}heorem \textbf{\underline{G}}eneration (ATG) benchmark that evaluates whether an agent can automatically generate valuable (and possibly brand new) theorems that are applicable for downstream theorem proving as reusable knowledge. 
Specifically, we construct the ATG benchmark by splitting the Metamath library into three sets: axioms, library, and problem based on their proving depth.
We conduct extensive experiments to investigate whether current LMs can generate theorems in the library and benefit the problem theorems proving. 
The results demonstrate that high-quality ATG data facilitates models’ performances on downstream ATP. 
However, there is still room for current LMs to develop better ATG and generate more advanced and human-like theorems. 
We hope the new ATG challenge can shed some light on advanced complex theorem proving. 
\end{abstract}

\section{Introduction}
\begin{figure}[htb]
  \centering
  \includegraphics[width=0.92\columnwidth]{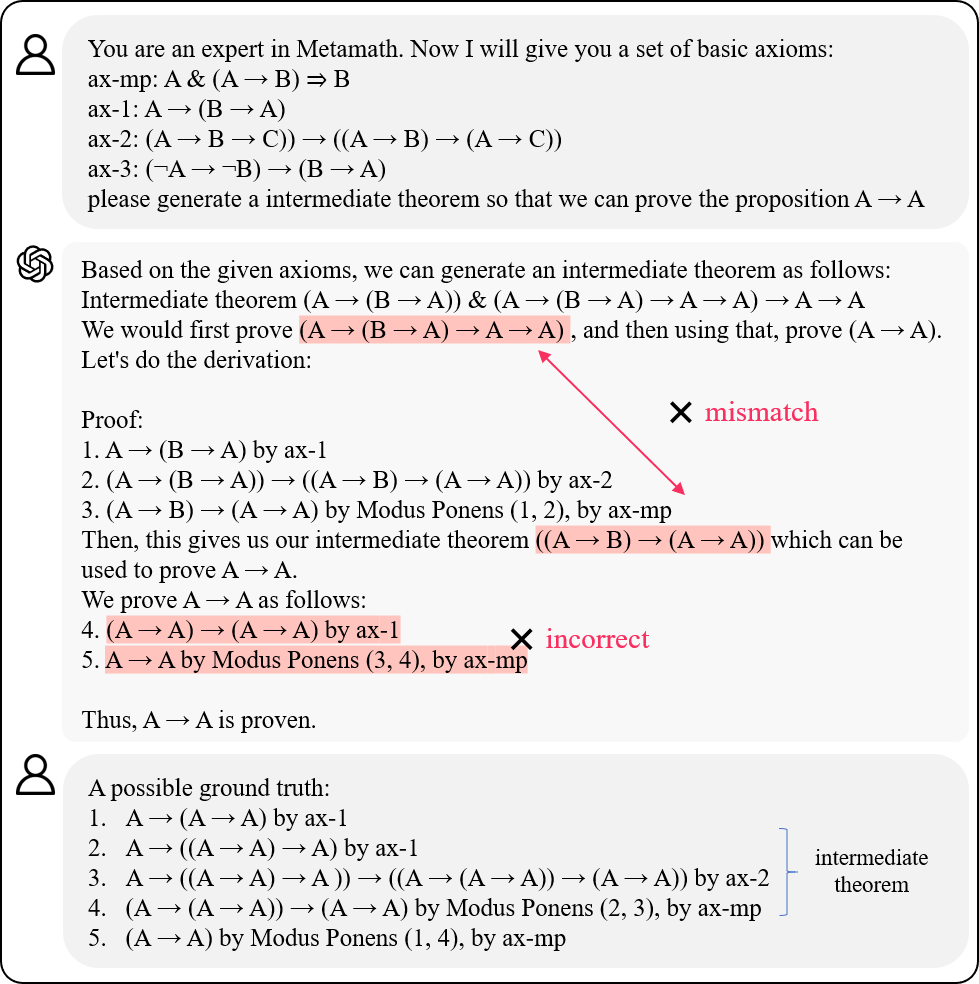}
  \caption{An example theorem generated 
  by GPT-4 \cite{gpt-4}. GPT-4 wrongly refers to the intermediate theorem $(A\rightarrow(B\rightarrow A)\rightarrow A \rightarrow A)$ as $((A \rightarrow B)\rightarrow(A\rightarrow A))$. 
  In Step 4, it applies ``ax-1'' but obtains the wrong expression instead of correct ($A\rightarrow(B\rightarrow A)$) 
  and can not derive ($A\rightarrow A$) even with the incorrect Steps 4 and 5.}
  \vspace{-5mm}
  \label{fig:GPT4}
\end{figure}

\begin{figure*}[!htb]
  \centering
  \includegraphics[width=0.92\textwidth]{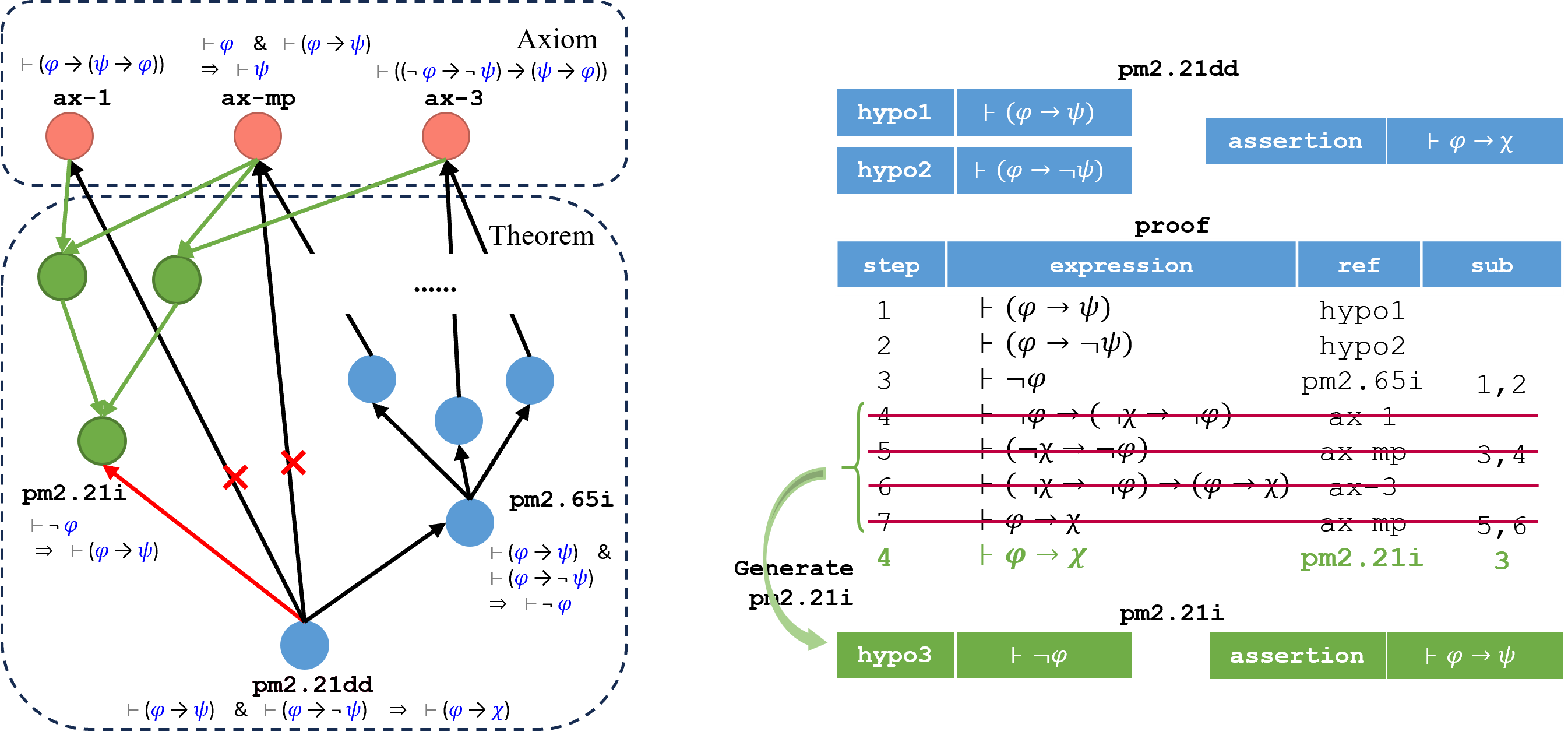}
  \caption{
  Illustration of 
  \yy{the Automated Theorem Generation (ATG) task and the process of proof reduction.}
  Black and green lines represent the proof and generation paths, while the red line is the new proof step. \modify{It takes 7 steps to prove the theorem ``pm2.21dd'' with given theorem ``pm2.21i'' and axioms while an ATG model can properly deduce the theorem ``pm2.21i'' and the total proof length of theorem ``pm2.21dd'' reduces from 7 steps to 4 steps.}
  }
  \label{fig:overview}
  \vspace{-7mm}
\end{figure*}

Recent generative language models (LMs) can perform advanced mathematical reasoning
%
including automated theorem proving (ATP) \cite{gpt-f} where the LMs need to provide a proof for a given theorem. 
Some studies \cite{first2023baldur} use the LMs to perform all-at-once generation to obtain the proof,
while another line of work \cite{gpt-f,han2021proof} leverages multi-step generation 
and combines reinforcement learning \cite{lample2022hypertree}, expert iteration \cite{polu2022formal, wang2023dt}, or reflection techniques \cite{yang2023leandojo} to simulate the multi-step search process.
%
Given the models achieve some complex theorem proving,
a shared limitation is their inability to reuse sub-propositions while proving or developing new theorems as humans do,
which also leads to redundant proving processes and low efficiency.
\yy{One demonstration is}
shown in Figure~\ref{fig:GPT4}.
Given a basic propositional logic system
\yy{(as shown in the top box)}, 
GPT-4 struggles deriving the basic proposition $(A \rightarrow A)$
\yy{(shown in the second box)}. 
\modify{Although \yy{GPT-4 being} the most advanced language model performs formal logical reasoning to a certain extent~\cite{yang2023leandojo}},
\yy{in this case,} \yy{it} wrongly refers to \yy{the} intermediate theorem in Step 3 and axioms in Steps 4 and 5. 
\yy{This case suggests an overlooked performance gap between LMs and humans in rigorous logical and structured derivation.}
\yy{Moreover,}
current state-of-the-art neural ATP methods~\cite{wu2022autoformalization, wang2023dt} do not reuse the searched sub-propositions in proving sequences. Thus they do not decompose the complex theorems into sub-problems and still suffer from the exponential search space.


\yynew{
To address the above-mentioned issues, 
we need to develop the capability of an agent to automatically create new and reusable theorems that are applicable for downstream theorem proving. 
For example, generating a theorem that can serve as a general sub-problem in proofs.
Moreover, such new theorems as a data source can further facilitate model parameter updates.
However, this question is under-explored and needs a clear problem definition and available data source.
}

This paper thus introduces the Automated Theorem Generation (ATG) task, 
\yy{where the LMs are required to automatically generate valuable theorems when given a set of axioms, and reduce the proof steps accordingly.}
\yy{Figure~\ref{fig:overview} illustrates the ATG task.}
\yy{In this case}, an ATG method 
\yy{is given axioms (``ax-1'', ``ax-mp'', and ``ax-3'') and should properly deduce the theorem ``pm2.21i'', which effectively simplifies the proof by reducing the original Steps 4-7 with the new Step 4.}
\yy{The proposed ATG task} has the following merits. 
Firstly, the forward deduction process \yy{in ATG} is more aligned \yy{with} text or code generation.
\yy{Since} the LMs have shown impressive performance~\cite{chen2021evaluating, roziere2023code}, 
\yy{the ATG task can better probe LMs' reasoning ability.}
Secondly, the generated theorems can be used in proving other theorems, and 
\yy{reduce} the \yy{proof} difficulties by 
\yy{deducing the intermediate steps from given axioms.}

\yy{Accordingly}, we construct \yy{an ATG} benchmark based on the Metamath formal system \cite{metamath} and its ``set.mm'' library\footnote{\href{https://github.com/metamath/set.mm/blob/develop/set.mm}{https://github.com/metamath/set.mm/blob/develop/set.mm}}. 
We also propose new metrics to evaluate the generated theorems by considering \yy{the} correctness, compactness, and usefulness of the proving target theorems.
\yy{Furthermore,} we combine the Monte Carlo tree search method~\cite{silver2017mastering, lample2022hypertree} with LMs and train \yy{the pipeline} with self-play policy learning to generate valuable theorems. 
\modify{The generated theorems, as augmented data training theorems prover, improved the performance of Holophrasm \cite{whalen2016holophrasm} and GPT-f \cite{gpt-f} by 16.16\% (passrate@1, from 20.48\% to 23.79\%) and 7.72\% (passrate@1, from 30.94\% to 33.53\%), respectively.}

\yy{Our contributions are summarized as follows:}
\vspace{-3mm}
\begin{itemize}
    \setlength{\itemsep}{-5pt}
    \setlength{\parsep}{-5pt}
    \setlength{\topsep}{-5pt}
    \item We 
    \yy{introduce}
    a new Automated Theorem Generation (ATG) task and propose a new dataset.
    \item We 
    \yy{define}
    \yy{a metric that specifically evaluates}
    the quality of \yy{ATG}
    which is independent of theorem provers. 
    \item We propose a self-play learning method and evaluate various methods on our proposed datasets to study the current state-of-the-art performance of theorem generation.
    \modify{\item We use the generated theorems as data augmentation and improve the performance of neural theorem provers.} 
\end{itemize}

\section{Related Work}
\vspace{-3mm}
\yy{Over the past few years, generative language models \cite{gpt-2,gpt-3,gpt-4} have achieved better mathematical reasoning. For example, solving math word problems \cite{dns,gsm8k,stepbystep}, 
linear programming \cite{nl4opt,he2022linear},
and parametric partial differential equations \cite{alet2019graph,anandkumar2019neural}.
Among the mathematical reasoning tasks, 
automated theorem proving \cite{irving2016deepmath,wang2017premise,selsam2018learning,alet2019graph,lachaux2020unsupervised,chaslot2008parallel} 
is one of the most challenging tasks as requires the neural models to perform consistent reasoning and rigorous multi-step deduction. 
More recent work \cite{gpt-f} applies the generative LM to automated theorem proving,
and following works improve the pipeline with kernel-level proof terms \cite{han2021proof}, expert iteration \cite{polu2022formal}, HyperTree Proof Search \cite{lample2022hypertree}, and sketching intermediate theorems \cite{DBLP:conf/iclr/JiangWZL0LJLW23}.
Such works suggest a great potential for generative LMs to prove theorems. 
However, a more advanced capability of synthesizing new and provable theorems and reusing them as intermediate proofs during proving remains unexplored. }

\yy{
Synthesizing theorems and their applications are in the preliminary research stage.
%
Previous works theoretically investigate theorem generation systems by executing computer programs \cite{johansson2009automated,sutcliffe2003grand,colton2001automated,lenat1977automated,lenat1976artificial, mccasland2006ascertaining,mccasland2006mathsaid} or by deriving from proof schemes \cite{buchberger2004algorithm,buchberger2006theorema,montano2012scheme}.
Such systems are barely applicable to modern neural models. 
Another line of work \cite{metagen,wu2020int,chou2000deductive,lample2022hypertree} leverages transformer-based LMs to generate theorems. 
However, 
the quality of the generated theorems is not guaranteed, and thus are less beneficial for downstream applications such as automated theorem proving (ATP). 
%
Therefore, this work proposes a rigorous task that challenges models to perform automated theorem generation and introduces corresponding metrics. 
To our knowledge, this is the first work on benchmarking neural automated theorem generation, and the resulting synthetic theorems can be directly applied to ATP. 
}


\begin{figure*}[t]
    \centering
    \includegraphics[width=\linewidth]{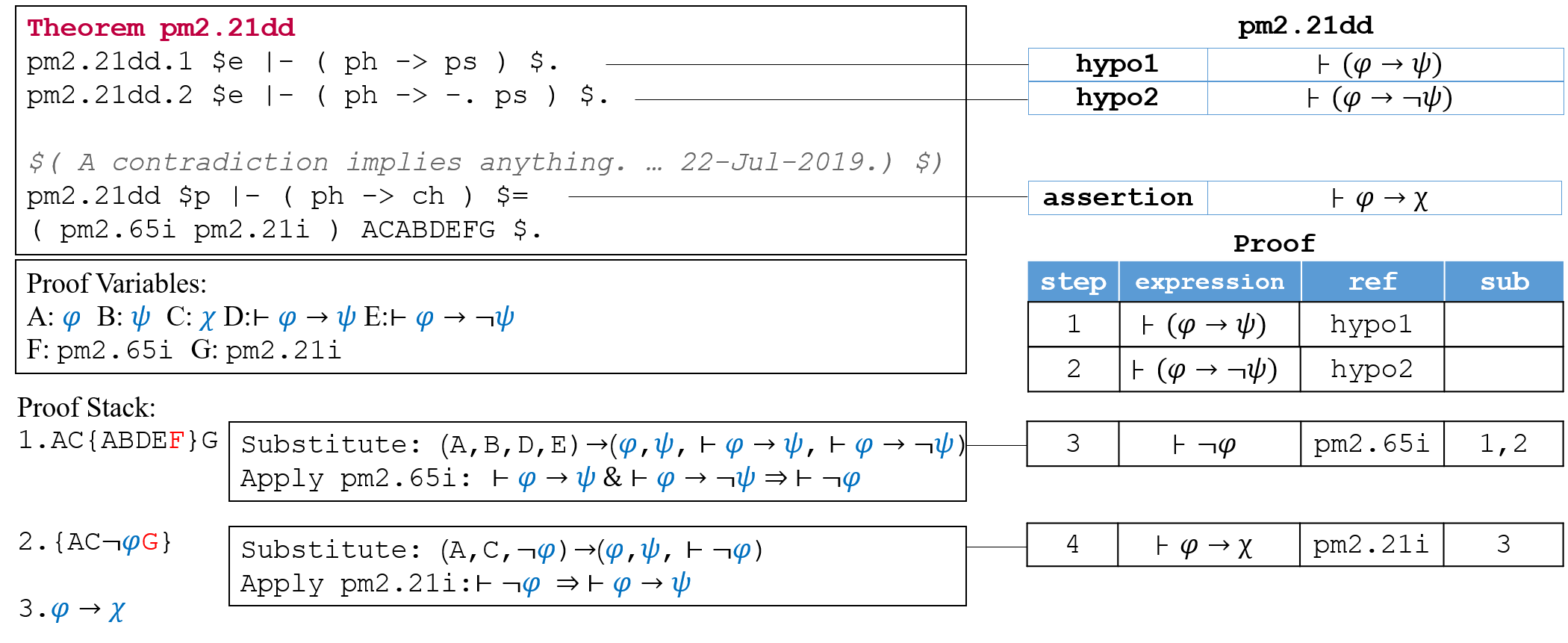}
    \caption{Example of proving with Metamath script. The script and verification process are shown on the left and the visualized proof step on the right. The script starts with defining symbols, hypotheses, and referred theorems. It pushes variables into the stack and applies substitution when encountering theorems. The applied result is pushed into the stack. The proof terminates when no more theorems are pushed in the stack.}
    \label{fig:code_example}
    \vspace{-5mm}
\end{figure*}

\section{\yy{Benchmarking Automated Theorem Generation}}


\vspace{-2mm}
\subsection{\yy{Preliminary with Metamath}}

\yy{Metamath \cite{metamath} is one of the widely used formal proof systems for proof verification and can interact with human and language models.}
\revise{It aims to describe rigorous mathematics with simple \textbf{substitution} operation.}

As shown in Figure~\ref{fig:code_example}, the upper left is an example Metamath script, \revise{which includes the statement and the proof (in a compressed form).}
\yy{The upper block shows two hypotheses (
\revise{denoted as ``pm2.21dd.1'' and pm2.21dd.2'' in line 2 and 3}
) and the assertion to be proved \revise{(\$p |- ( ph -> ch ) \$=, denoted as ``pm2.21dd'' in line 6)}.
\revise{In line 7,} it also shows the referred theorems \revise{(the names of the theorems in parentheses)} and the compressed proof sequence ``ACABDEFG''. 
The lower block gives the assignments \revise{of these uppercase letters}. 
}


\revise{The Metamath compiler treats the proof sequence as a list of operations on the proof stack. The statements represented by these capital letters perform push/pop operations on the proof stack according to their type.}
When first encountering letter ``F'', since ``F:pm2.65i'' 
\revise{is a theorem which}
has four variables (two for symbols and two for hypotheses), the Metamath program pops four elements (``ABDE'') from the stack, substitutes the ``pm2.65i'' variables with (``ABDE''), returns the proved expression $\neg\psi$ and pushes it to the stack. 
Metamath program performs a similar operation when encountering the other referred theorems ``G:pm2.21i''. It substitutes $\varphi, \psi, \neg\psi$ with first $3$ stack elements ($\varphi, \chi, \neg\psi$), respectively, and then returns the substituted assertion $\varphi \rightarrow \chi$. 
\yy{The proof is regarded as complete once there is only one element remaining in the stack and it is equal to the assertion. }
\revise{Appendix~\ref{app:detailed_preliminary} describes the detailed procedure for this proof.}

\vspace{-2mm}
\subsection{Task Definition}
\yy{We introduce the task of \textbf{A}utomated \textbf{T}heorem \textbf{G}eneration (ATG), as we assume that valuable theorems should be applicable for further proving, while it is intuitive that it is effortless for one to generate numerous and random theorems.}

\yy{The task is formally defined as follows.}
Given an initial theorem library $\mathbb{L}=\{T_1, T_2, \cdots, T_n\}$ 
\yy{that consists} of axioms $T_i$, and a problem set $\mathbb{P}=\{P_1, P_2, \cdots, P_m\}$ where the problem $P_i$ has the shortest proof $p_i = \{t_1, t_2, \cdots, t_n \vert t_i\in \mathbb{L}\}$ deduced from the initial library $\mathbb{L}$, 
\yy{an automated theorem generation algorithm}
$G$ is required to expand the initial library $\mathbb{L}$ to $\mathbb{L'} = \{T_1, T_2, \cdots, T_n, T_{n+1}, \cdots, T_{n+k}\}$ with new theorems $\mathbb{L}_G=\{T_{n+1}, \cdots, T_{n+k}\}$, such that for problem $P_i \in \mathbb{P}$, its shortest proof $p_i' = \{t_1', t_2', \cdots, t_n' \vert t_i'\in \mathbb{L'}\}$ is shorter than $p_i$.


\vspace{-2mm}
\subsection{Evaluation Metrics}

\paragraph{Average Proof Reduction}
We propose average proof reduction ($APR$) to evaluate the generation quality.
\modify{
\yy{The assumption is}
that shorter proofs are preferred, \revise{\cite{2024arXiv240304571B}}
or else the search space of the proof will grow exponentially \yy{as} the proof length \yy{increases}, thus affecting the performance of theorem proving.
Besides, The proof length of a theorem 
\yy{indicates its level of abstraction.}}

\yy{The $APR$ metric \revise{automatically} measures} the number of reduced proof steps given the expanded theorem library $\mathbb{L'}$. We first define proof distance $D(\mathbb{L},\mathbb{P}$) as the average proof length from theorem library $\mathbb{L}$ to problem set $\mathbb{P}$:
\begin{equation}
\label{for:proof distance}
\setlength\abovedisplayskip{3pt}
\setlength\belowdisplayskip{3pt}
    D(\mathbb{L},\mathbb{P}) = \frac{1}{m} \sum_{i=1}^{m} len(p_i), 
\end{equation}
where $p_i$ is the shortest proof of problem $P_i \in \mathbb{P}$ deduced from the library $\mathbb{L}$. 
Recall that all problems $P_i$ has \yy{its shortest} proof $p_i$ given \yy{the} initial $\mathbb{L}$ and $\mathbb{L}\subseteq\mathbb{L'}$, thus problem $P_i$ is always provable with $\mathbb{L'}$ and $D(\mathbb{L},\mathbb{P})$ has a feasible value. 
We then define the \textbf{average proof reduction ($APR$)} as:
\begin{equation}
\label{for:APR}
\setlength\abovedisplayskip{3pt}
\setlength\belowdisplayskip{3pt}
    APR = D(\mathbb{L},\mathbb{P}) - D(\mathbb{L}',\mathbb{P}) - len(\mathbb{L}_G),
\end{equation}
\yy{where the constraint term $len(\mathbb{L}_G)$ restricts the number of generated theorems $\mathbb{L}_G$. }
\yy{It avoids the models from generating} too many theorems
that will 
increase the searching difficulty in \yy{downstream} automated theorem proving.

\yy{To distinguish whether the generated theorem is helpful or irrelevant for theorem proving, we introduce two evaluation criteria as follows:}
\vspace{-3mm}
\begin{enumerate}
\itemsep0em 
\item 
\yy{We} check whether the generated theorem 
\yy{matches} any theorem in the ground truth ``set.mm'' library. 
\yy{This criterion ensures that}
the generated theorem is \yy{included} in the original proof and thus certainly helps the proof. 
\item
We \yy{inspect} if both the hypotheses and the assertion \yy{in} the generated theorem
\yy{are consistent with the} intermediate results \yy{in} the ground-truth proof. 
\yy{This second criterion guarantees that the generated theorem is a proper substitution of the original proof segments.}
\end{enumerate}
\vspace{-2mm}
For instance, in Figure~\ref{fig:overview}, the hypothesis $\neg\phi$ \yy{in} theorem ``pm2.21i'' matches the intermediate Step 3, and we \yy{thus} obtain the assertion $\phi \rightarrow \chi$ that matches Step 7. 
\yy{Therefore,}
the generated theorem ``pm2.21i'' 
\yy{helps reduce}
Steps 4 - 7.
$APR$ is independent of theorem provers. 
It \yy{only} depends on the axiom system and the amount of information in the theorem library.
\yy{As a result, this metric can be generally applied to ATG evaluation.}

\vspace{-2mm}
\paragraph{Human-Aligned Precision}
\yy{We consider the human-written theorems in the ``set.mm'' library as the standard and anticipate that the generated theorems should align with human-written ones.}
\yy{Thus we compute the}
precision of generated theorems set $\mathbb{L}_G$ over ``set.mm'' $\mathbb{L}_h$: 
\begin{equation}
\label{for:precision}
\setlength\abovedisplayskip{3pt}
\setlength\belowdisplayskip{3pt}
\yy{\text{Precision} =} \ \  len(\mathbb{L}_G \cup \mathbb{L}_h) / len(\mathbb{L}_G).
\end{equation}

\vspace{-3mm}
\paragraph{\revise{Theorem Count}}
\yy{To further investigate the extent to which a model expands the initial theorem library and generates how many theorems, we also evaluate the \revise{theorem count}.
}
\yy{The measure is the direct} $len(\mathbb{L}_G)$.





\begin{figure}[t]
  \centering
  \includegraphics[width=1.0\columnwidth]{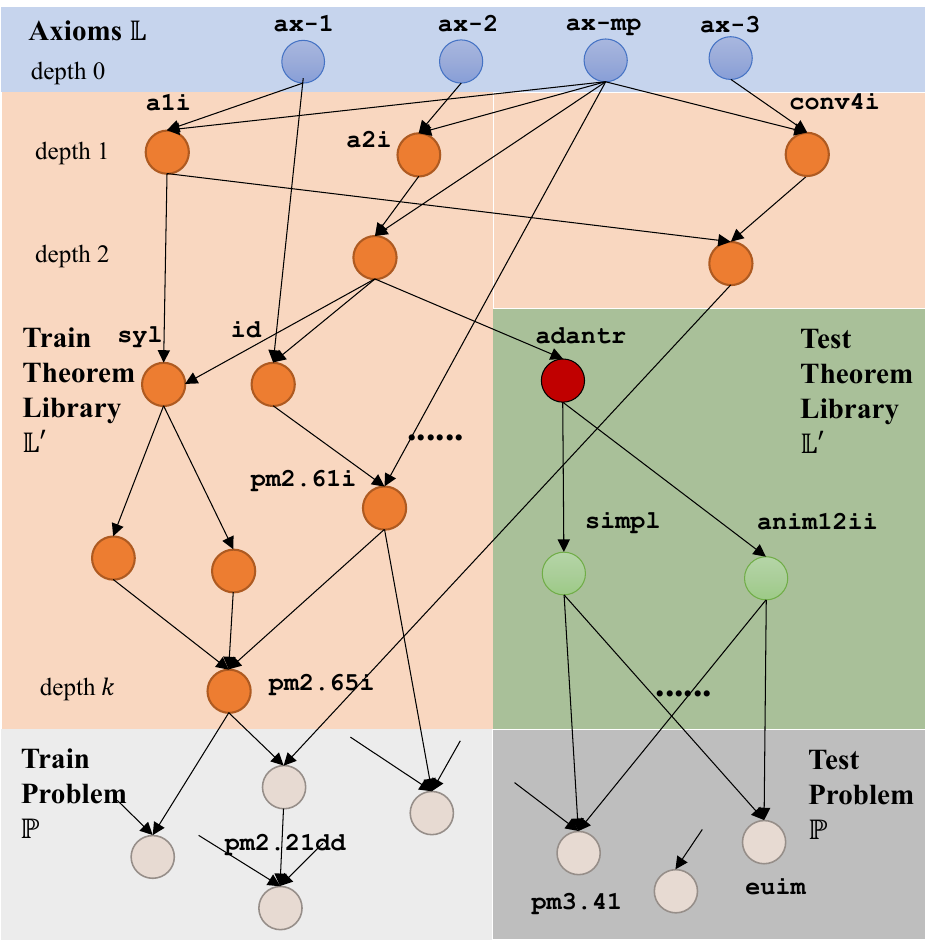}
  \caption{
  The construction process of the ATG benchmark. Each node is a theorem in ``set.mm'' and edges represent if a node refers to another in its proof. We assign each node a depth and use it to split the theorem library $\mathbb{L'}$ and $\mathbb{P}$. Lastly, we select the red node and use all its successor nodes for testing.
  }
  \label{fig:dataset}
  \vspace{-5mm}
\end{figure}

\vspace{-1mm}
\subsection{Dataset Construction}
We construct the ATG benchmark based on the Metamath formal language \cite{metamath} and its ``set.mm'' set. The ``set.mm'' library comprises $\sim 38k$ human-written theorem\revise{s}. To create \revise{ a dataset that consists of concise, fundamental, and high-quality theorems while encompassing a broad range of theorems,} we focus on a subset of around 2,000 theorems specifically related to propositional calculus. To evaluate the reasoning ability of a generation method $G$ at different levels, we further construct ``wb'', ``wif'' and ``minimp'', $3$ datasets with different complexity. We construct a directed acyclic graph of theorems for each dataset, and split the theorem library and problem set according to the depth of the theorem in the graph. As shown in Figure~\ref{fig:code_example}, a theorem proof in "set.mm" involves references to hypotheses and other proven theorems. Using the reference relations, we create a directed graph that illustrates how a theorem is deduced from the axioms.
As demonstrated in Figure~\ref{fig:dataset}, we represent each ``set.mm'' theorem as a node, then draw an edge from referenced theorem A to B to indicate B is proved with A. The resulting directed acyclic reference graph encompasses all theorems connected to the axioms. The depth of each theorem node is assigned based on its longest distance to the axiom nodes. For example, the axiom ``ax-1'' has depth $0$, and theorems ``a1i'', ``syl'' have depth $1$ and $3$. 

We build the initial theorem library $\mathbb{L}$ with axioms. 
Then, theorems with depth less or equal to $k$ are assigned to theorem library $\mathbb{L'}$, and the others are assigned to problem set $\mathbb{P}$.
Lastly, we split the training and testing theorems mostly randomly with balanced sample numbers such that the training and test set have minimum overlap library $\mathbb{L'}$, problem set $\mathbb{P}$, and corresponding proof path $p$.
More details on constructing the dataset are described in~\ref{app:construction}.

\begin{table}[t]
  \centering
  \resizebox{\linewidth}{!}{
  \begin{tabular}{*{8}{c}}
    \toprule
            & Axioms \&  & Theo- & Set  & Split & Theorem  & Problem \\
    Dataset & Hypotheses & rems  & Type & Depth & Library  & Set \\
    \midrule
    \multirow{2}{*}{wb}  & \multirow{2}{*}{83}  & \multirow{2}{*}{272} & train & 10 & 82  & 32  \\
                             &                      &  & test  & 20 & 54  & 21  \\
    \multirow{2}{*}{wif}  & \multirow{2}{*}{247} & \multirow{2}{*}{1284} & train & 33 & 518 & 220 \\
                             &                      & & test  & 39 & 211 & 88  \\
    \multirow{2}{*}{minimp} & \multirow{2}{*}{373} & \multirow{2}{*}{2048} & train & 36 & 754 & 298 \\
                             &                      & & test  & 40 & 441 & 182 \\
    \bottomrule
  \end{tabular}}
  \vspace{-3mm}
  \caption{Statistics of proposed ATG datasets.}
  \label{tab1}
  \vspace{-4mm}
\end{table}

\begin{figure}[t]
    \centering
    \includegraphics[width=\linewidth]{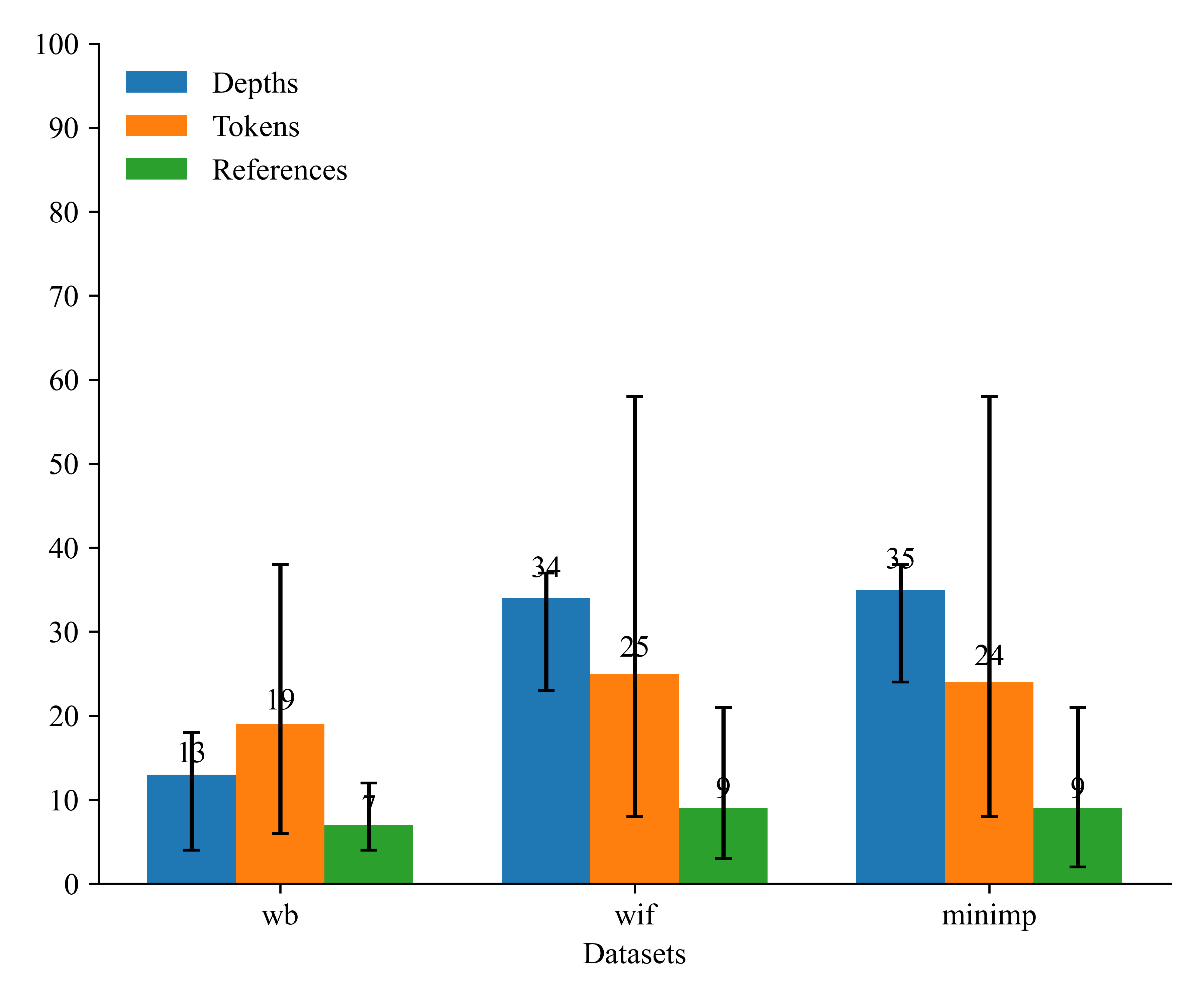}
    \vspace{-5mm}
    \caption{
    Statistics of proof depth, tokens, and referred theorems of test theorem library.}
    \label{fig:testL}
    \vspace{-6mm}
\end{figure}

\begin{figure*}[t]
    \includegraphics[width=\linewidth]{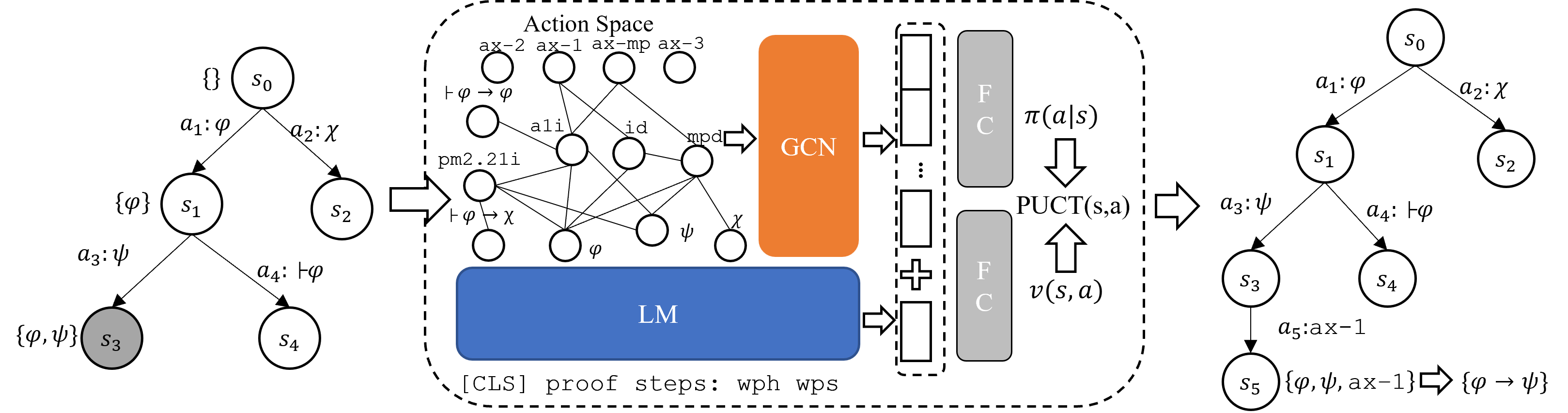}
    \caption{Overview of single-step expansion by our proposed method MCTS+pvn. We use the proof stack as state and applicable axioms or hypotheses as action. We use a language model and graph network to encode states and actions for obtaining PUCT scores. State-action pair with the highest PUCT is expanded to the next state.}
    \label{fig:framework}
    \vspace{-5mm}
\end{figure*}

\subsection{Dataset Statistic}
\yy{As demonstrated in Table~\ref{tab1}, }
the ``wb'' dataset contains 83 axioms and hypotheses, and ``wif'' and ``minimp'' dataset contains 247 and 373, respectively. 
More statistics of the test theorem library in Figure~\ref{fig:testL} show a large variance across all three datasets.
Theorems in the ``wb'' dataset have a relatively low depth \yy{(max=18, min=4, average=13)},
\yy{while ``wif'' and ``minimp'' have higher depths (``wif'' average=34 and ``minimp'' average=35).} 
\revise{On the other hand,}
the token numbers of the statement of the theorem do not differ much across the three datasets, indicating similar difficulty for LMs to generate a single theorem.
\yy{Moreover,} the number of references is $9$ for both the ``wif'' and ``minimp'' and $7$ for ``wb''.
More details of the dataset statistical information are shown in Appendix~\ref{Dataset Information}.

\vspace{-2mm}
\section{Self-Play Policy Learning}
\vspace{-3mm}
In this section, we propose a theorem generation method based on Monte Carlo tree search and self-play learning. 
We use the current proof stack as a state and act to apply one possible axiom, hypothesis, or theorem. Starting with an empty state, we iteratively select a state and action to expand based on \textbf{p}olicy/\textbf{v}alue \textbf{n}etwork (\textbf{pvn}) output until a valid proof or maximum length is reached.
During training, we sample state-action/state-value pairs with policy/value networks and use the sample pairs as supervised signals.

\begin{table*}[t]
  \centering
  \resizebox{\linewidth}{!}{
  \begin{tabular}{llll|lll|lll}
    \toprule
    \multirow{2}{*}{Methods} & \multicolumn{3}{c}{wb}  & \multicolumn{3}{c}{wif} & \multicolumn{3}{c}{minimp}   \\
                             & $len(\mathbb{L}_G)$  & $APR$$\uparrow$ & Precision$\uparrow$ & $len(\mathbb{L}_G)$  & $APR$$\uparrow$ & Precision$\uparrow$  & $len(\mathbb{L}_G)$ & $APR$$\uparrow$  & Precision$\uparrow$ \\
    \hline
    BPE                      & 66   & 21.41  & 15.79      & 504   & 532.49 & 44.26     & 697   & 662.23 & 46.88 \\
    \hline
    Random                   & 129  & 1.66   & 0.00       & 407   & 32.67  & 0.00      & 136   & 61.21  & 0.00  \\
    MCTS                     & 3384 & 29.98  & 0.00       & 22585 & 456.04 & 0.00      & 28870 & 530.73 & 0.00  \\
    MCTS+pvn                 & 3697 & 34.56  & 1.85       & 30184 & 482.81 & 1.18      & 35798 & 548.89 & 0.77  \\   
    \bottomrule
  \end{tabular}}
  \caption{Performance for BPE, random search, MCTS, and MCTS+PVN methods on wb, wif, minimp datasets.}
  \label{tab2}
  \vspace{-5mm}
\end{table*}

\paragraph{Action} Actions are axioms or hypotheses that can be applied during theorem generation. We use a subset of possible actions during each generation. Specifically, the sub-action space includes all axioms and all symbols such as $\varphi, \psi, \chi$. We randomly add $5$ theorems from the current generated theorem library $\mathbb{L}'_t$ and then sample $5$ hypotheses from proof of the selected theorems. We further build a graph indicating their referring relation, as shown in Figure~\ref{fig:framework}.

\vspace{-3mm}
\paragraph{State} The state represents the current proof stack. As shown in Figure~\ref{fig:framework}, we start with an empty state ``\{\}'' $s_0$. If we apply action $\varphi$, then the resulting state $s_1$ is ``{$\varphi$}''. Further applying action $\psi$ and ``ax-1'', we obtain the state $s_5=\{\varphi,\psi,\text{ax-1}\}$ and reduce to ${\varphi\rightarrow\psi}$.

\vspace{-3mm}
\paragraph{Monte Carlo Tree Search}
We expand the initial empty state iteratively until we reach a valid proof or a maximum step. Figure~\ref{fig:framework} shows the process of one-step expansion. At each step, we first obtain the policy probability and value for all state-action pairs. \yy{Specifically,} we apply GPT2-large \cite{gpt-2} to encode the action history with special prefix tokens. For example, to encode state $s_3$, we input sequence ``[CLS] PROOFSTEP wph wps'' to the LM, where ``[CLS] PROOFSTEP'' is the prefix and ``wph'' and ``wps'' represent $\phi$ and $\psi$ in Metamath. We also use a graph convolution network \cite{kipf2016semi} to encode the action and their relations. The action features are \yy{then} concatenated with the state feature, then flattened and fed into a fully connected layer to obtain \modify{the value of current state $v(s,a)$ and the policy probability $\pi(a\vert s)$ for all possible actions.} 
Given the policy $\pi(a\vert s)$ and the value $v(s,a)$, we obtain the PUCT \cite{silver2017mastering} scores:
\begin{equation}
\label{for:puct}
\small
\setlength\abovedisplayskip{3pt}
\setlength\belowdisplayskip{3pt}
PUCT(s, a) = v(s, a) + c\pi(s,a) \frac{\sqrt{\sum_b N(s,b)}}{1+N(s,a)},
\end{equation}
where $c$ is a factor to balance exploration and exploitation and $N(s,a)$ is the visit count for state-action pairs ($s,a$). $N(s,a)$ adds $1$ if its descendant nodes are expanded at current step.

\vspace{-3mm}
\paragraph{Reward and Theorem Generation} The search stops once it reaches a valid proof or a maximum step. We give a state reward $1$ if it is a new theorem $T$ or $0$ otherwise. At each episode $i$, we repeat the search process with the same sub-action space, and select the most value theorem $T_i$ for theorem library expansion:
\label{for:update library}
\setlength\abovedisplayskip{3pt}
\setlength\belowdisplayskip{3pt}
\begin{align}
T_i &= \arg\min_{T_i} D(\mathbb{L}_i \cup \{T_i\}, \mathbb{P}_{train}), \nonumber \\
\mathbb{L}_{i}' &\leftarrow \mathbb{L}_{i-1}' \cup \{T_i\}, 
\end{align}
where $\mathbb{L}'_{0}=\mathbb{L}$ is the the initial theorem library. We stop expand $\mathbb{L}_{i}'$ if the model in episode $i$ does not generate any new theorem.

\begin{figure}[t]
  \centering
  \includegraphics[width=0.95\columnwidth]{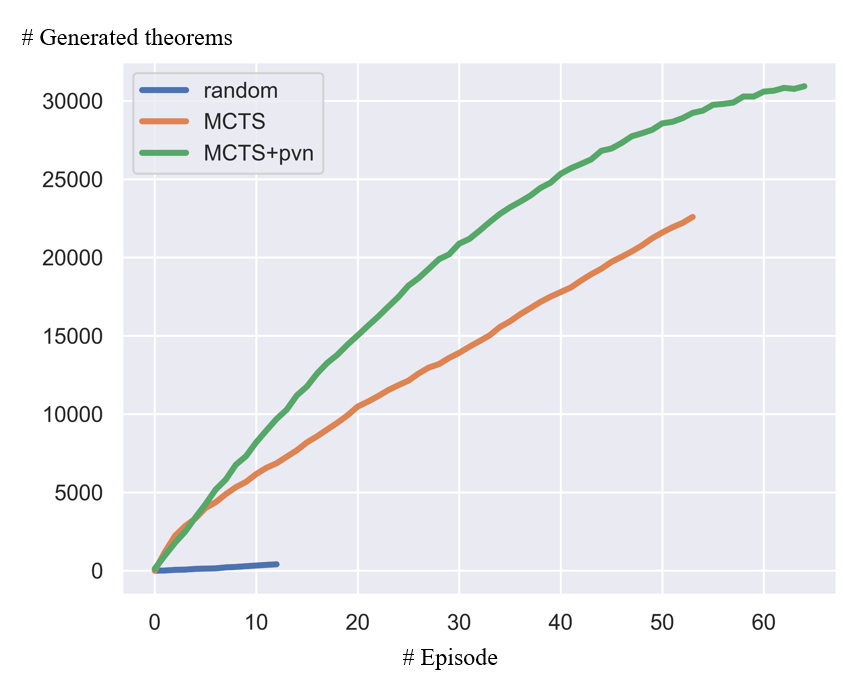}
  \vspace{-2mm}
  \caption{Number of generated theorems across generation episodes.}
  \label{generate-episode curve}
  \vspace{-5mm}
\end{figure}

\begin{figure}[t]
    \centering
    \includegraphics[width=0.95\columnwidth]{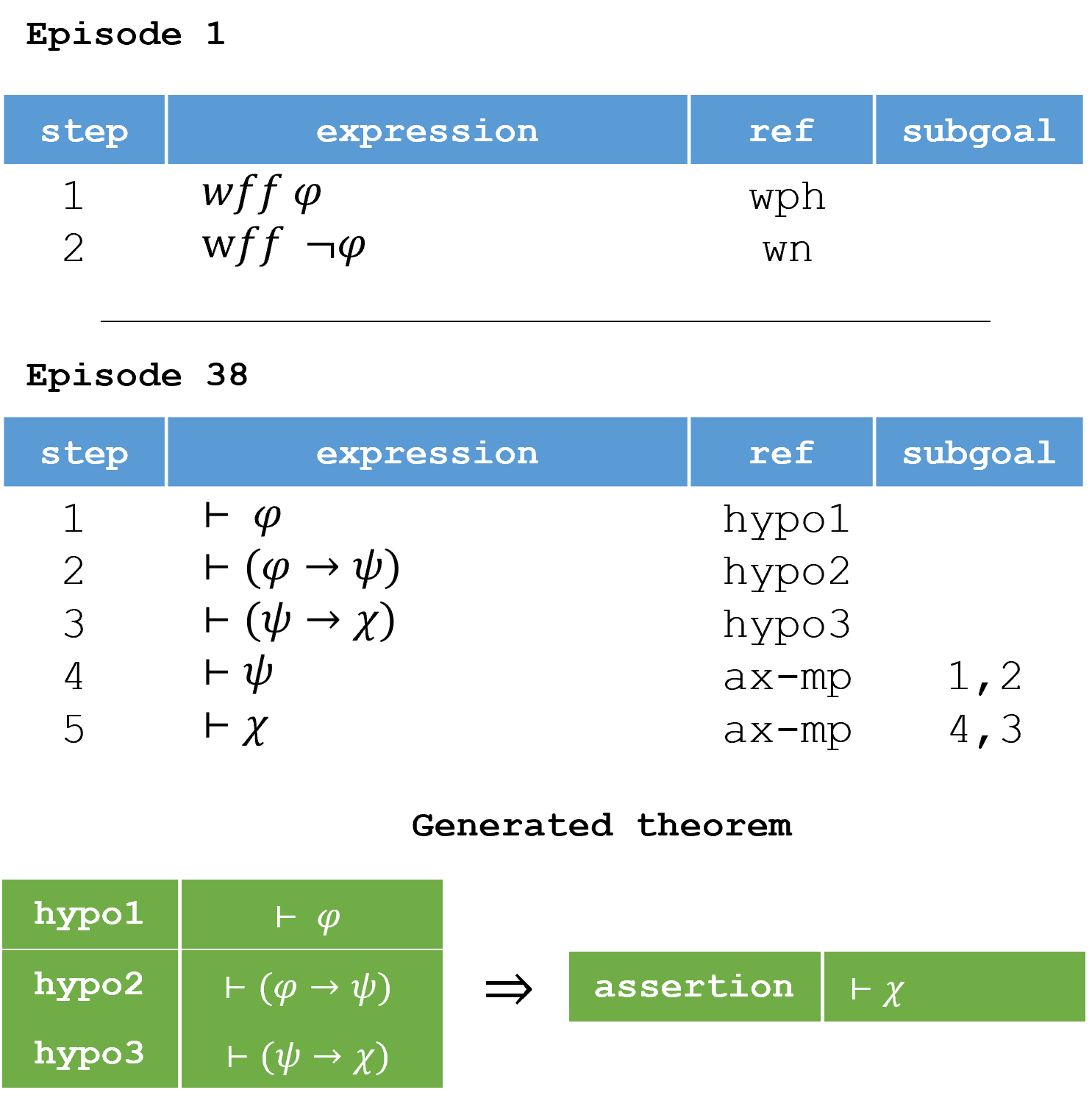}
    \caption{Example theorems by \textbf{MCTS+pvn}. 
    This is the exact theorem ``mp2b'' in the ``set.mm'' library.}
    \label{fig:generate_samples}
    \vspace{-5mm}
\end{figure}

\vspace{-3mm}
\paragraph{Self-Play Learning}
We use its own policy and value networks to perform theorem generation to assist training. During each training iteration, we first perform theorem generation $100$ times with current policy/value network parameters. For all search results, we back-propagate the reward to the ancestors with a discount factor $\gamma$ to obtain (state, value) pairs. We also count the action frequencies of all traveled states and obtain the (state, action) probabilities. We also use the theorems in training library $\mathbb{L}'$ as successful search results and obtain (state, action) (state, value) in the same way. All obtained (state, action) are used to train the policy network with KL divergence loss, and (state, value) are used to train the value network with MSE loss.

\revise{
\vspace{-3mm}
\paragraph{Inference}
During inference, the algorithm starts from an empty stack and uses a neural network for predicting the $\pi(a\vert s)$ and $v(s,a)$. Then the $PUCT$
 value is calculated using Equation~\ref{for:puct}. We then perform MCTS to select the optimal node and execute the corresponding action. We use Metamath rules to determine whether a new theorem is generated: If so, update the theorem library according to Equation~\ref{for:update library}, and enter the iteration of the next episode; If no, regenerate the theorem from an empty stack.
}

\begin{table*}[t]
    \small
    \centering
        \begin{tabular}{lccccccccc}
            \toprule
            $c_{puct}$ & 0.0 & 0.1 & 0.3 & 0.5 & 1 & 3 & 5 & 10 & 100 \\ 
            \midrule
            $len(\mathbb{L}_G)$ & 11645 & 24883 & 30184 & 27921 & 26325 & 25583 & 22847 & 20453 & 16840 \\ 
            APR$\uparrow$ & 304.27 & 413.85 & 482.81 & 452.65 & 407.31 & 395.42 & 375.83 & 334.33 & 335.18 \\ 
            Precision$\uparrow$ & 0.36 & 0.89 & 1.18 & 1.03 & 0.91 & 0.91 & 0.82 & 0.75 & 0.63 \\ 
            \bottomrule
        \end{tabular}
    \caption{Performance of different $c_{puct}$ values.}
    \label{tab:ablation:c_puct}
    \vspace{-5mm}
\end{table*}

\section{Experiments}
\vspace{-2mm}
\subsection{Baseline Methods}
In addition to our proposed \textbf{MCTS+pvn} trained with self-play, we also evaluate three baseline methods:
The random search policy, traditional Monte Carlo tree search (MCTS), and a Byte Pair Encoding (BPE) based statistical method.
Note that BPE retrieves theorems from the human-written library $\mathbb{L'}$ as an approximation of human performance.
\yy{Details of baseline models are introduced in Appendix~\ref{app:baseline} and the details of implementation are demonstrated in Appendix~\ref{app:implementation}.}

\vspace{-1mm}
\subsection{\yy{Main Results}}
\yy{The compared theorem generation abilities are demonstrated in Table~\ref{tab2}. We have the following observations:}
(1) Random search policy suffers from poor performance as it lacks any form of intelligent decision-making, generating only a few basic theorems and they do not overlap with the human theorem library. 
(2) MCTS policy gradually improves its search efficiency and can make informed choices by leveraging the information gathered from previous simulations. 
(3) \textbf{MCTS+pvn} helps to improve search efficiency and decision quality by incorporating learned knowledge and heuristics into the exploration and exploitation process. 
In general, 
compared with the human-approaching BPE method, current methods show significant performance gaps and still struggle to synthesize theorems as humans.

\begin{table}[t]
    \centering 
    \resizebox{\linewidth}{!}{
    \begin{tabular}{lcc}
    \toprule
    Training data  & Prover     & Pass (\%) \\
    \midrule
    set.mm  & Holophrasm & 20.48    \\
    set.mm + MetaGen  & Holophrasm & 22.06    \\
    set.mm + ours & Holophrasm & 23.79    \\
    set.mm + MetaGen + ours & Holophrasm & 24.30    \\
    set.mm              & GPT-f      & 30.94         \\
    set.mm + augmented  & GPT-f      & 31.58    \\
    set.mm + ours  & GPT-f      & 33.53   \\
    \bottomrule
    \end{tabular}
    }
\caption{Generated theorems as augmented data improve ATP performances.}
\vspace{-3mm}
\label{tab:ATP}
\end{table}

\vspace{-1mm}
\subsection{Generation Efficiency}
Given infinite time, a generative method $G$ can eventually obtain as many provable theorems.
\yy{However, we prefer a $G$ that generates more valid theorems with fewer generation episodes. }
Specifically, we observe how many theorems can be generated by the search policies except for the BPE method. 
The results are shown in Figure~\ref{generate-episode curve}. 
As the search space complexity grows,
it becomes more difficult for search policies to generate new theorems, which further reflects the challenge of our dataset. The random policy ends generation early with few generated theorems, while heuristic search methods generate new theorems at high speed. The generation efficiency of the MCTS with a neural network is inferior to pure MCTS without training, but finally achieves better generation performance with learning from a combination of human-written data and self-play data.

\vspace{-2mm}
\subsection{Case Study}
Figure~\ref{fig:generate_samples} shows the theorems generated by 
\textbf{MCTS+pvn}
on the ``minimp'' dataset. In the first few episodes, our algorithm only attempts to construct simple expressions using symbols and axioms that conform to Metamath syntax. In episode 38, with the introduction of proper hypotheses, the algorithm successfully applies the axiom ``ax-mp'' twice and deduces a provable conclusion. The result shows that the policy benefits from self-play learning and learning to reason in the formal system. More examples of generated theorems are attached in~\ref{app:generation examples}.

\begin{table}[t]
  \small
  \centering
  \begin{tabular}{llcc}
    \toprule
    Model       & params & APR     & delta  \\
    \midrule
    BERT-base   & 104M & 407.85  &  \\
    BERT-large  & 320M & 430.38  & +22.53 \\
    \midrule
    GPT2        & 124M & 397.21  &  \\
    GPT2-medium & 355M & 411.73  & +14.52 \\
    GPT2-large  & 744M & 482.81  & +85.60  \\
    \midrule
    OPT-350m    & 350M & 491.89  &  \\
    OPT-1.3b    & 1.3B & 495.47  & +3.58 \\
    \midrule
    Llama 2-7B  & 7B   & 507.71  &  \\
    Llama 2-13B & 13B  & 512.30  & +4.59  \\
    \midrule
    Mixtral-7B  & 7B   & 511.64  &  \\
    Mixtral-8x7B & 45B & 527.29  & +15.65 \\
    \bottomrule
  \end{tabular}
  \caption{``wif'' results with different LM scales.}
  \label{tab:ablation:model}
  \vspace{-5mm}
\end{table}

\vspace{-2mm}

\subsection{\revise{Benifits for Theorem Proving}}

\begin{table*}[t]
    \small
    \centering
        \begin{tabular}{lcccccccc}
            \toprule
            \# problems help to solve & 1$\sim$2 & 3 & 4 & 5 & 6 & 7 & 8 & $\geq 9$ \\ 
            \midrule
            \# theorems & 22831 & 3346 & 1532 & 734 & 327 & 217 & 16   & 1181 \\
            \bottomrule
        \end{tabular}
    \caption{The distribution of the number of problems solved by the generated theorems.}
    \label{tab:n_helps_solve}
    \vspace{-5mm}
\end{table*}

\revise{To gain a more comprehensive understanding, we conducted additional analysis on theorems generated by our MSTS+pvn method on the "wif" dataset. We evaluate the number of "wif" test problems whose proof has utilized the generated theorems. The results are shown in Table~\ref{tab:n_helps_solve} and show that our method can produce generalized theorems that closely resemble axioms and solve multiple problems. These generalized theorems involve declarative premises, basic definition inferences, and similar elements. Some of these theorems are illustrated in Appendix~\ref{app:generation examples}.}

We further evaluate whether the generated theorems are helpful for automated theorem proving. 
The results are shown in Table~\ref{tab:ATP}. 
We test two provers: the Holophrasm~\cite{whalen2016holophrasm} prover and the GPT-f~\cite{gpt-f} prover. 
The implementation details are demonstrated in Appendix~\ref{app:impl_atp}.

We observed that after co-training with the generated ATG theorems (ours), the Holophrasm prover achieves an improvement of 90 generated theorems (i.e., 20.48\% $\to$ 23.79\% in the Holophrasm test set).
Moreover, co-training including the ATG and 
MetaGen~\cite{metagen} theorems,
the number of theorems improves by 104 (i.e., 20.48\% $\to$ 24.30\% in the Holophrasm test set). 
%
Furthermore, GPT-f~\cite{gpt-f} synthesizes some theorems about n-digit arithmetic and ring algebra, and the pass rate is improved to 31.58\%. With our ATG theorems, the pass rate is further improved to 33.53\%.
Therefore, the generated high-quality data are suggested as beneficial to automated theorem proving. 

\vspace{-2mm}
\subsection{Ablation Study}
We ablate different model variants and see the performances.
We first explore whether the model scale affects the results. 
We evaluate some open-source language models~\cite{devlin2018bert, gpt-2, zhang2022opt} on the ``wif'' dataset. The result in Table~\ref{tab:ablation:model} demonstrates that 
the language model scale has a remarkable influence on the generation. 
Models with more parameters achieve a higher $APR$ on the same dataset. Besides, we only use the language model to encode the proof steps, and the decoder-only models do not show the same advantages on ATG task as it does on text generation tasks. 

\modify{
We also investigate the impact of different $c_{puct}$ in Equation (\ref{for:puct}) on search performance. 
When $c$ approaches infinity, PUCT approaches breath-first search.
While $c$ approaches 0, PUCT becomes a depth-first search. 
The results of \textbf{MCTS+pvn} with different $c_{puct}$ are
demonstrated in Table~\ref{tab:ablation:c_puct}. 
It is suggested that when $c$ is 0.3, the algorithm reaches the balance between exploration and exploitation and achieves the best performance. 
}

\vspace{-2mm}
\section{Conclusion} 
\vspace{-2mm}
This paper introduces the Automated Theorem Generation (ATG) task that evaluates generative language models' capability of generating high-quality theorems and reducing complex theorems. 
We also introduce evaluation metrics that consider the correctness, compactness, and usefulness of the generated theorems.
We conduct extensive experiments and find that high-quality human-like theorems benefit downstream tasks such as automated theorem proving. 
Moreover, there is still room for current language models to generate such high-quality theorems. 
Therefore, we hope the proposed ATG benchmark can facilitate the development of language models' theorem generation and thus improve the overall theorem proving. 

\section*{Limitations}
This paper introduces a new Automated Theorem Generation (ATG) task and a corresponding benchmark.
There are several directions for future research:
\begin{itemize}
    \item We only construct the training and test problems set based on theorems depth in the ``set.mm'' library. However, an ideal problem set should be general enough to represent the distribution of all of the math problems in the real world.
    \item We sample several hypotheses randomly from the training library during each generation process. Developing a method to automatically generate non-contradictory hypotheses remains a challenge.
    \item We only build the ATG dataset on the Metamath language. A more general ATG benchmark should include other formal systems such as Lean and Isabelle.
\end{itemize}

\section*{Ethics Statement}
The proposed ATG datasets are constructed based on the open-sourced human labeled ``set.mm''  library.
No personal or confidential information is collected.
Therefore, to the best of our knowledge, there is no ethical concern.

\section*{Acknowledgements}
This work was supported in part by  
National Key R\&D Program of China under Grant No. 2020AAA0109700,  
Guangdong Outstanding Youth Fund (Grant No. 2021B1515020061),
Mobility Grant Award under Grant No. M-0461, 
Shenzhen Science and Technology Program (Grant No. RCYX20200714114642083), 
Shenzhen Science and Technology Program (Grant No. GJHZ20220913142600001), 
Nansha Key RD Program under Grant No.2022ZD014, the China Postdoctoral Science Foundation under Grant Number 2023M744001.

\bibliography{ref}

\appendix
\section{Appendix}
\subsection{Detailed proving process of theorem ``pm2.21dd''}
\label{app:detailed_preliminary}
1. In Metamath, a \textbf{variable} is denoted as a Greek letter (e.g., $\phi$,$\psi$,$\chi$, or the identical "ph", "ps", "ch" in Figure~\ref{fig:code_example}), representing a mathematical object or concept. "$\rightarrow$" denotes \textbf{entailment} between the variables, and "-." denotes \textbf{negation} of the variables. The symbol "|-" means the following symbol sequence is \textbf{provable} or a proof exists for it. For example, "|- (ph -> -.ps)" in Figure~\ref{fig:code_example} indicates that $\phi$ (indicated by "ph") yields not (indicated by "-.") $\psi$ (indicated by "ps"), and it's provable.

2. A \textbf{hypothesis} is denoted as "[hypothesis name] \$e [hypothesis] \$." For example, in Figure~\ref{fig:code_example}, "pm2.21dd.1" and "pm2.21dd.2" are two essential hypotheses. Similarly, an \textbf{assertion} to be proved is denoted as "[assertion name] \$p [assertion] \$." In Figure~\ref{fig:code_example}, the conclusion of the theorem pm2.21dd is "(ph -> ch)". 

3. A \textbf{theorem} in general consists of several hypotheses (sometimes none) and one conclusion. In the case in Figure~\ref{fig:code_example}, the theorem "pm2.21dd" has two hypotheses "pm2.21dd.1" and "pm2.21dd.2" and one assertion to be proved "pm2.21dd". The theorem can be stated in natural language as follows: if $\phi$ deduces both $\psi$ and $\neg \psi$, then it deduces $\chi$. In other words, a contradiction implies anything.

4. Figure~\ref{fig:code_example} show the \textbf{proving process} of the theorem "pm2.21dd" in the Metamath system. In Figure~\ref{fig:code_example}, the symbol "\$=" in the assertion statement is followed by the proof of this theorem.

5. In the following line, "pm2.65i" and "pm2.21i" in parentheses are the axioms or proven theorems that will be referred to the proof. They are later denoted by F and G in the proof sequence, respectively. In this case, variables F and G are regarded as \textbf{operators} in the calculator, while variables A to E are \textbf{operands} in the calculator. The sequence "ACABDEFG" is the inverse Polish expression of a proof. 

6. The assignment of operands and operators has the following rules: The symbol showing up first in the theorem statement is assigned first. In this case, A, B, and C denote $\phi$, $\psi$, and $\chi$. After that are the hypotheses: D denotes "pm2.21dd.1", and E denotes "pm2.21dd.2". Finally are the references: F for "pm2.65i" and G for "pm2.21i".

7. The prover \textbf{program} handles the proof process with the inverse Polish expression of the proof sequence. This means that it maintains \textbf{a proof stack} (initially empty) and uses operands and operators to push and pop the stack in turn. As shown in ~\ref{fig:code_example}, since the first 6 elements are operands, the program simply pushes them into the stack. The seventh element (red F shown in ~\ref{fig:code_example}) is "pm2.65i", which has four operators by its statement (2 symbols $\phi$ and $\psi$, and 2 hypotheses "pm2.65i.1" and "pm2.65i.2"). Therefore, the program pops four elements from the stack, replaces the corresponding symbols and hypotheses in the original statement ("\textbf{Substitute}:" in ~\ref{fig:code_example}), gets the conclusion $\vdash \neg \phi$ ("\textbf{Apply}:" in ~\ref{fig:code_example}), and pushes it into the stack (proof step 3 in ~\ref{fig:code_example}). This process shows the only basic rule "\textbf{substitution}" in Metamath.
8. Next, the program pushes the eighth element "pm2.21i" denoted by the red G, which has three operators. The program processes it in a similar process and pushes the resulting $\phi \rightarrow \psi$ into the stack (proof step 4 in ~\ref{fig:code_example}). At this point, the proof sequence is processed, and the only remaining elements in the stack agree with the conclusion of the theorem, which means that the theorem is successfully proved.

In Figure~\ref{fig:framework}, the right part shows a brief proof of the "pm2.21dd" theorem. The first and second lines of proof refer to two hypotheses of this theorem, "pm2.21dd.1" and "pm2.21dd.2", respectively. In the third line of the proof, the theorem "pm2.65i" is referenced and the expression is shown in the left graph. "pm2.21dd.1" and "pm2.21dd.2" are used respectively to replace the original hypotheses "pm2.65i.1" and "pm2.65i.2" of the theorem "pm2.65i", and derive the conclusion $\vdash \neg \phi$. That is, if $\phi$ deduces both $\psi$ and $\neg \psi$, then $\phi$ is a false statement. The proofs in the following lines 4-7 show the process of deriving the conclusion $\vdash \phi \rightarrow \chi$ from the original set of axioms, which can be summarized as the theorem "pm2.21i", that false propositions derive everything.
This case can also be seen HERE \footnote{\href{https://us.metamath.org/mpeuni/pm2.21dd.html}{https://us.metamath.org/mpeuni/pm2.21dd.html}} for more information.

\subsection{Details of Dataset Construction}
\label{app:construction}

We construct the ``wb'' set by selecting the first $272$ theorems and then split the theorem library $\mathbb{L'}$ and problem set $\mathbb{P}$ at depth $k=10$ and $k=20$ for training and testing. Similarly, we construct the ``wif'' and ``minimp'' sets with $1,284$ and $2,048$ theorems, respectively. The problem set has depths larger than $k=33$, $k=38$ for training, and $k=39$, $k=40$ for testing.

\subsection{Dataset Information}
\label{Dataset Information}
\begin{figure*}[htbp]
    \centering
    \begin{subfigure}{\columnwidth}
    \includegraphics[width=\textwidth]{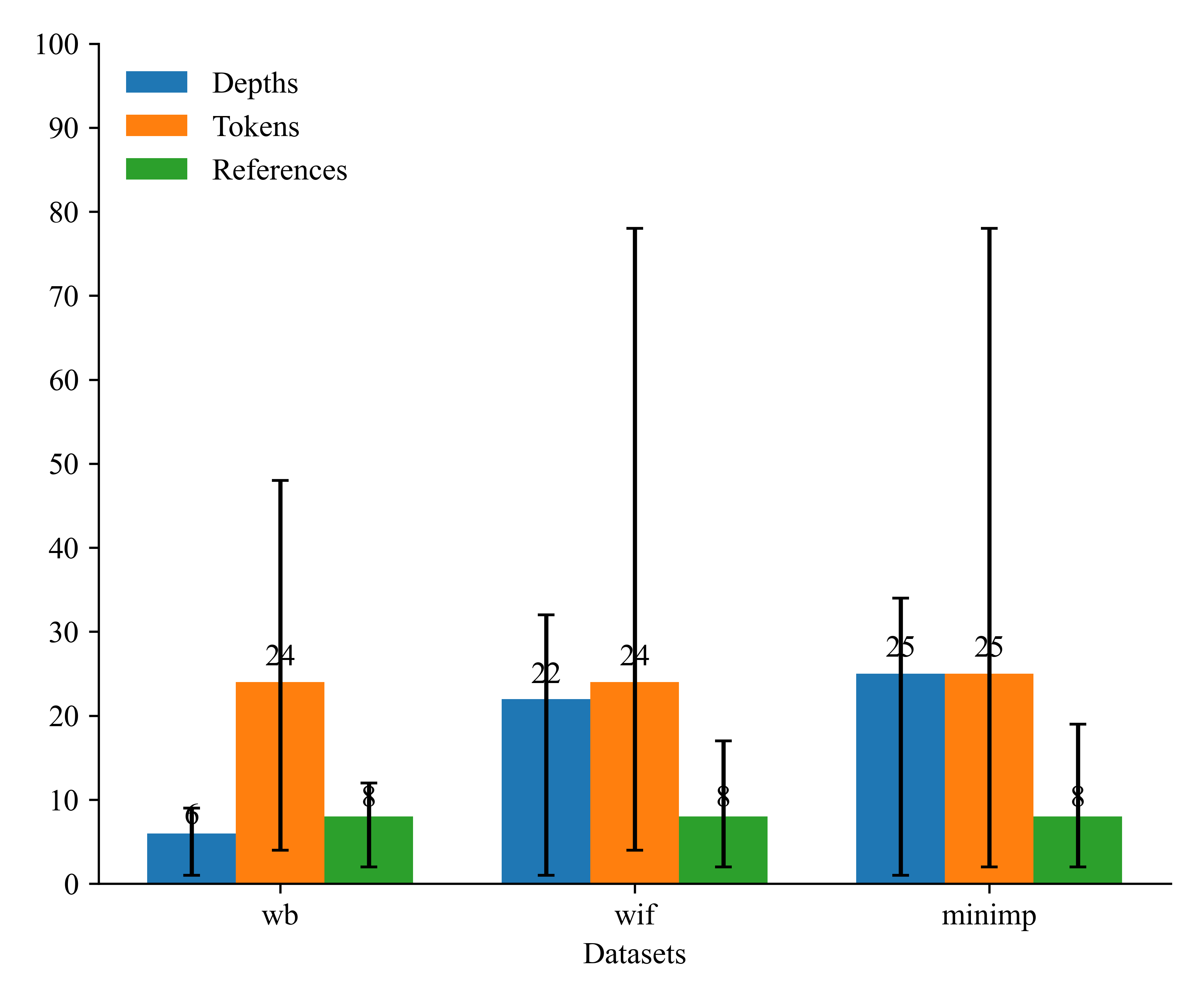}
    \caption{Statistical information of the train theorem library.}
    \end{subfigure}
    \begin{subfigure}{\columnwidth}
    \includegraphics[width=\textwidth]{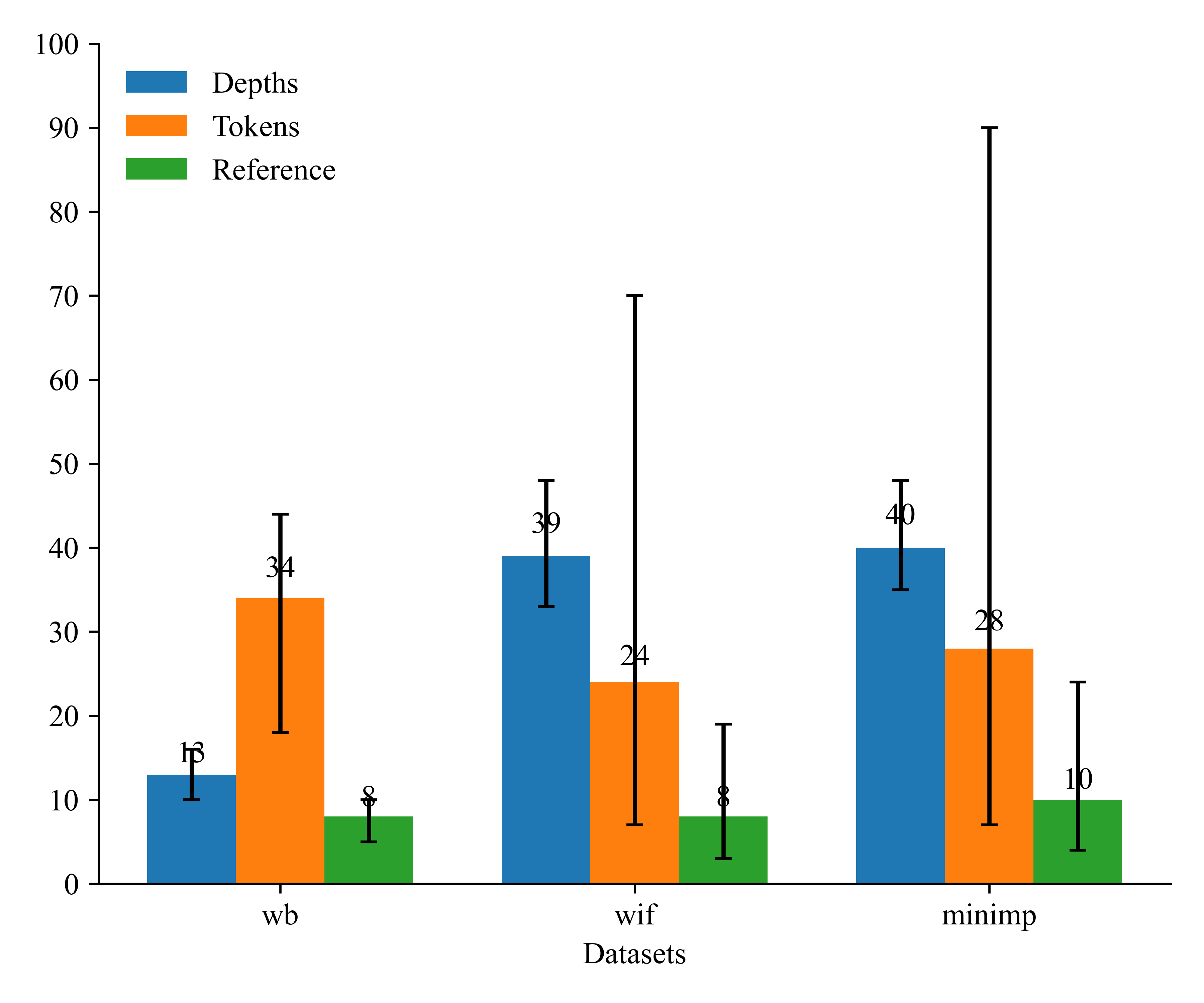}
    \caption{Statistical information of the train problem set.}
    \end{subfigure}
    \begin{subfigure}{\columnwidth}
    \includegraphics[width=\textwidth]{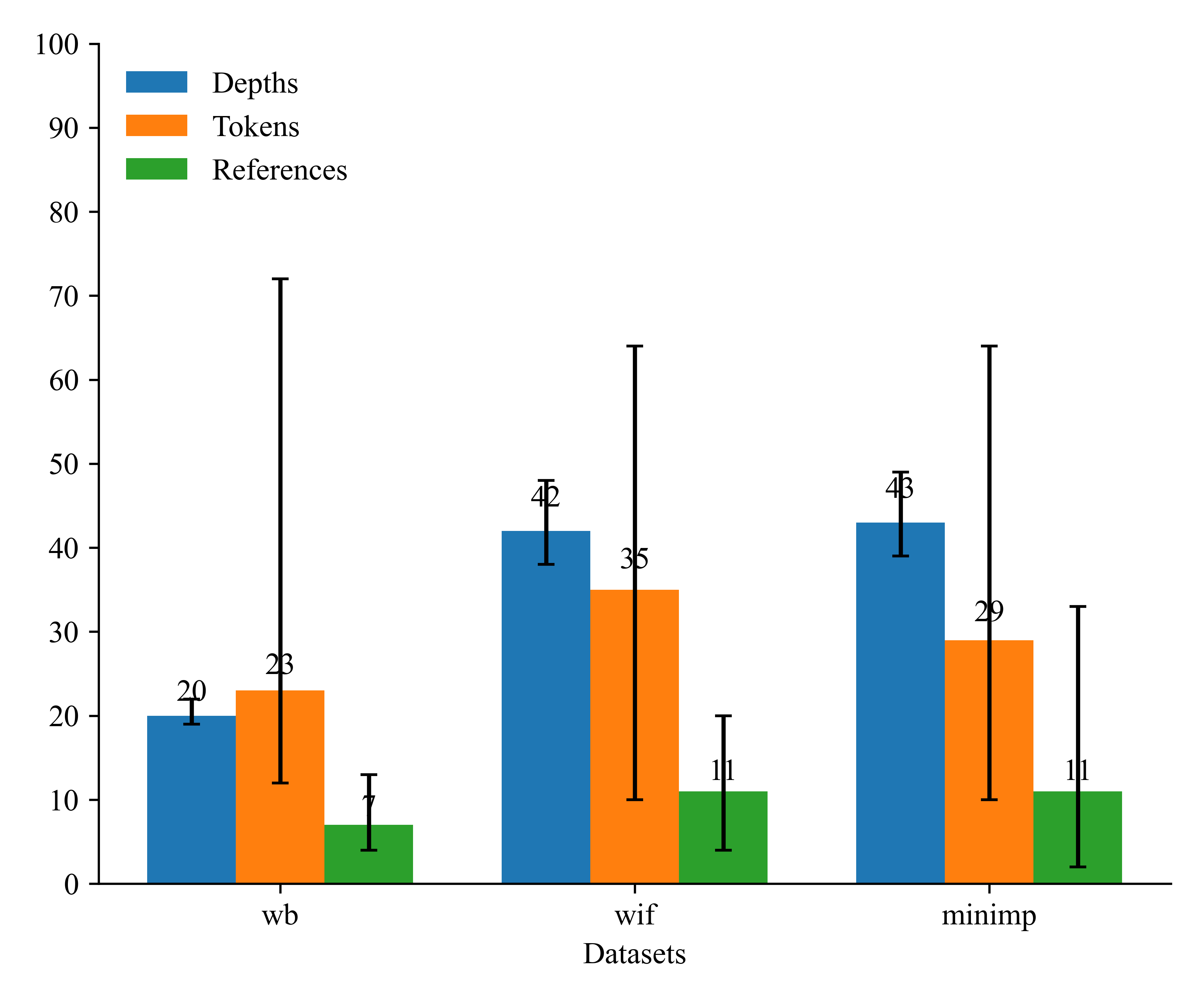}
    \caption{Statistical information of the test theorem library.}
    \end{subfigure}
    \begin{subfigure}{\columnwidth}
    \includegraphics[width=\textwidth]{fig/test_theorem_library.png}
    \caption{Statistical information of the test problem set.}
    \end{subfigure}
    \caption{The average, min, max numbers of proof Depth, tokens and referred theorems of all three datasets.}
    \label{fig:dataset visualize}
\end{figure*}

The details of the statistical information of all three datasets are listed in Table~\ref{tab:Dataset_Details}. The resulting visualization is shown in Figure~\ref{fig:dataset visualize}.

\begin{table*}
  \centering
  \begin{tabular}{l*{11}{c}}
    \toprule
    \multirow{2}{*}{Dataset} & \multicolumn{3}{c}{Depth} & \multicolumn{3}{c}{Token} & \multicolumn{3}{c}{References} \\
        & min & avg & max & min & avg & max & min & avg & max \\
    \midrule
    wb train library & 1  & 6 & 9 & 4 & 24 & 48 & 2 & 8 & 12 \\
    wb train problem & 10 & 13 & 16 & 18 & 34 & 44 & 5 & 8 & 10 \\
    wb test library  & 4  & 13 & 18 & 6 & 19 & 38 & 4 & 7 & 12 \\
    wb test problem  & 19 & 20 & 22 & 12 & 23 & 72 & 4 & 7 & 13 \\
    \midrule
    wif train library & 1  & 22 & 32 & 4 & 24 & 78 & 2 & 8 & 17 \\
    wif train problem & 33 & 39 & 48 & 7 & 24 & 70 & 3 & 8 & 19 \\
    wif test library  & 23 & 34 & 37 & 8 & 25 & 58 & 3 & 9 & 21 \\
    wif test problem  & 38 & 42 & 48 & 10 & 35 & 64 & 4 & 11 & 20 \\
    \midrule
    minimp train library & 1  & 25 & 34 & 2 & 25 & 78 & 2 & 8 & 19 \\
    minimp train problem & 35 & 40 & 48 & 7 & 28 & 90 & 4 & 10 & 24 \\
    minimp test library  & 24 & 35 & 38 & 8 & 24 & 58 & 2 & 9 & 21 \\
    minimp test problem  & 39 & 43 & 49 & 10 & 29 & 64 & 2 & 11 & 33 \\
    \bottomrule
  \end{tabular}
  \caption{Detailed statistics of wb, wif, minimp datasets in the ATG benchmark.}
  \label{tab:Dataset_Details}
\end{table*}

\subsection{Baseline Methods}
\label{app:baseline}
In addition to our proposed \textbf{MCTS+pvn} trained with self-play, we also evaluate three baseline methods. The random search policy, traditional Monte Carlo tree search (MCTS), and a Byte Pair Encoding (BPE) based statistical method to find theorems given the proof in training theorem library $\mathbb{L'}$.

\vspace{-1mm}
\paragraph{Random Search} We explore the search space by expanding the most recent state with randomly selected actions, without considering their potential outcomes or evaluating their quality.

\vspace{-1mm}
\paragraph{Monte Carlo Tree Search} \origin{MCTS applies Upper Confidence Bounds (UCB) algorithm to select expanded states and actions without explicit policy probability. To obtain the value for each state-action pair, MCTS perform simulations that randomly walk to a terminal state, collect rewards and propagate back to the state-action pair and update all nodes' value in this path.}
\modify{We perform random walks in simulation to select the expanded nodes rather than employing a policy-value network. Specifically, we first expand the unexplored sub-nodes. If all the sub-nodes have been explored, we select the next node to expand according to the PUCT score in Equation~(\ref{for:puct}) in Section 4, where $\pi(a\vert s)$ of all available actions are set to the same, and $v(s,a)$ is determined by the visits times of the node. During each node expansion, we obtain the node value by randomly expanding the deepest node until it reaches a terminal state.  The terminal state will return a reward which is backporpated to all nodes in this path and update their value.}

\vspace{-1mm}
\paragraph{Byte Pair Encoding} 
Byte Pair Encoding (BPE) is an unsupervised subword tokenization algorithm used in natural language processing. It starts with a vocabulary of characters and subwords and iteratively merges the most frequent character or subword pairs in the corpus, updating the vocabulary accordingly. In ATG, we construct vocabulary from the human-written theorems in training library $\mathbb{L'}$, \origin{each token in the proof text refers to the name of an axiom, hypothesis, or theorem, and a proof ends with a special token $[EOS]$. 
To obtain a valid subword,
if the BPE-merged pair has an axiom or a theorem as the second item, we include previous tokens in the proof text and expand the merged subword until it is a valid proof.
We locate all such pair in all training proof text and consider all resulting subwords as a generated theorem $T$.
The algorithm terminates when the frequency of all token pairs is 1.}
\modify{Specifically, we use the full name of axioms, hypotheses, and theorems as the minimum elements (token) for grouping. 
Then we select the most frequent pairs in all training proofs and merge them into a new element (subword). 
Then we replace this pair with the new elements in all training proofs and perform merging iteratively until all pair occurrence is 1.
To obtain a valid subword, if the merged pair has an axiom or a theorem as the second item, we include previous tokens in the proof and expand the merged subword until it is a valid proof. Note that the BPE method is a statistical method rather than a search algorithm, and that the source text is derived from human-written proofs, we employ it as an approximation of human-level performance.}

\subsection{Implementation Details of ATG}
\label{app:implementation}
We evaluate all of our baseline approaches with a maximum proof step length of 32 and generate 100 times in each episode, the generation ends until no more valuable theorem is generated. All of the generations begin from an initial graph that only consists of axioms and hypotheses in corresponding datasets. Methods with MCTS simulate 100 times for expanding the search tree and take a $c$ value $0.3$ to balance exploration and exploitation. We collect $(state, action, value)$ triplets from proofs in the human written library $L'$ and combine data generated by MCTS policy to train the policy/value network in each episode. The training process contains 10 epochs per episode and takes 128 samples per batch. The learning rate is set to $3e-4$ and we train our models with the Adam optimizer. We also apply an early-stop mechanism at minimum KL divergence between policy probabilities before and after training.

\subsection{Implementation Details of ATP}
\label{app:impl_atp}
We test two provers: the Holophrasm prover and the GPT-f prover. 
Because different provers use different versions of ``set.mm'' libraries, 
the data pre-processing and the training pipeline vary accordingly.
The generated theorems are then formatted into standard Metamath language form as augmented data.
We further use the BPE method to generate more theorems based on the
``minimp'' dataset, and then randomly substitute the variables in the theorems to construct an augmented theorem sets with
$\sim 30k$ proof steps.

During evaluation, we follow the original Holophrasm prover \cite{whalen2016holophrasm} and the original GPT-f prover \cite{gpt-f} setups and use the same data splits. 


\subsection{More Examples of Generated Theorems}
\label{app:generation examples}
\begin{figure*}[htbp]
    \centering
    \includegraphics[width=\linewidth]{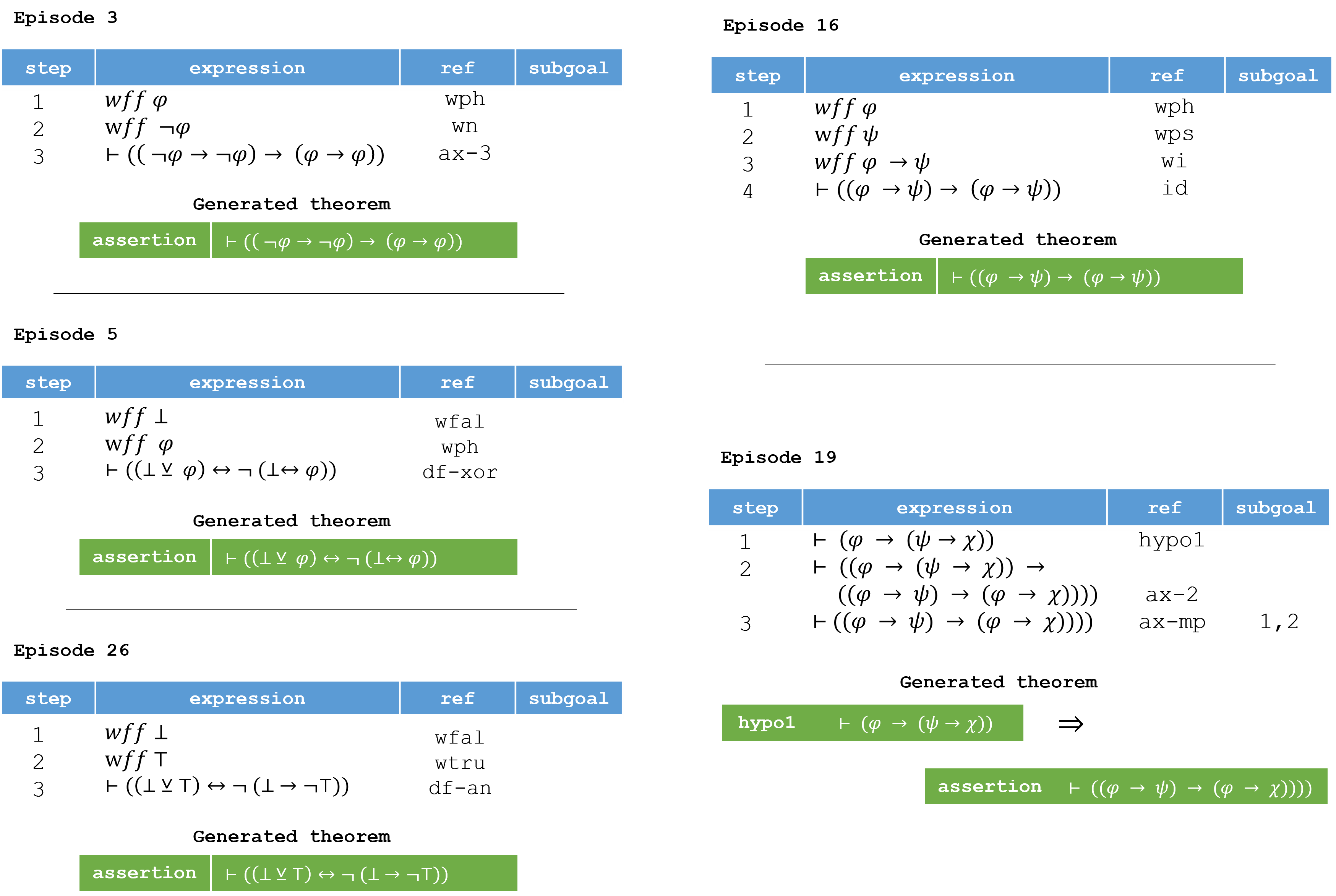}
    \caption{Some generated theorems are referred to in \yy{downstream} ATP tasks.}
    \label{fig:generate samples}
\end{figure*}

\begin{figure}[t]
    \centering
    \includegraphics[width=0.95\columnwidth]{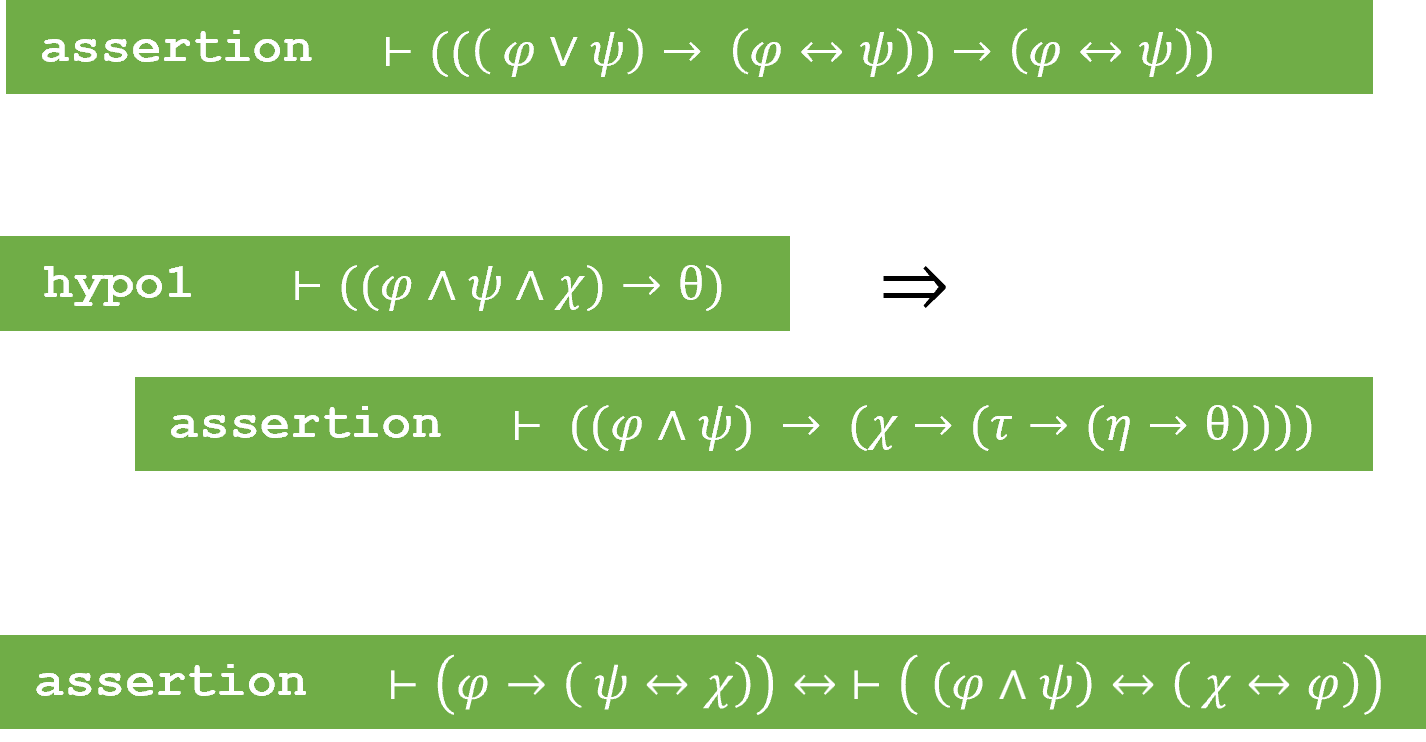}
    \caption{Example complex theorems generated by the proposed MCTS+pvn.}
    \label{fig:generate_samples_hard}
    \vspace{-5mm}
\end{figure}

Figure~\ref{fig:generate samples} provides more theorems generated by MCTS with policy/value network on the ``minimp'' dataset, these theorems are referred when proving other theorems in downstream ATP tasks. Figure~\ref{fig:generate_samples_hard} shows more complex theorems generated by our method. These theorems involve complex concepts in propositional logic such as the distribution of implication over biconditional and require as many as dozens of proof steps.


\end{document}